\documentclass[letterpaper,10pt,conference]{ieeeconf}

\usepackage{times}
\usepackage[pdftex]{graphicx}
\usepackage{amsmath,amssymb}
\usepackage{bm}
\usepackage[space]{cite}
\usepackage{array}
\usepackage{booktabs}
\usepackage[skip=3pt,font={small}]{caption}
\usepackage[linkcolor=black,citecolor=black,urlcolor=black,colorlinks=true]{hyperref}

\DeclareGraphicsExtensions{.png,.jpg,.jpeg,.pdf}
\IEEEoverridecommandlockouts
\title{\LARGE \bf DuctAM: A Duct-Assisted Quadrotor-Based Aerial Manipulator Enabling High-Force Push-and-Pull Interactions
}

\author{ Yi Wang$^*$, Rui Jin$^*$, Xinhang Xu$^*$, Haotian Jin, Ruiyang Liu, Yizhuo Yang and Lihua Xie$^\dagger$
    \thanks{ $^*$Indicates equal contribution. }
	\thanks{$^\dagger$Corresponding author: {\tt\small elhxie@ntu.edu.sg}.}
	\thanks{
	Yi Wang, Rui Jin, Xinhang Xu, Haotian Jin, Ruiyang Liu, and Yizhuo Yang are with the School of Electrical and Electronic Engineering, Nanyang Technological University, Singapore 639798.
	Lihua Xie is with the NTU--VinUni Joint Research Laboratory for Embodied AI and Robotics, School of Electrical and Electronic Engineering, Nanyang Technological University, Singapore 639798, and VinUniversity, Hanoi, Vietnam.
	}
    \thanks{
    This work was supported by Ministry of Education, Singapore, under AcRF TIER 1 Grant RG64/23.
	}
}

\begin{document}

\maketitle
\thispagestyle{empty}
\pagestyle{empty}

\begin{abstract}
Uncrewed Aerial Manipulators (UAMs) extend the capabilities of Uncrewed Aerial Vehicles (UAVs) from perception to physical interaction. Among various aerial interactions, push-and-pull operations are fundamental manipulation primitives that require sustained horizontal forces while maintaining stable flight. In this paper, we propose DuctAM, a compact aerial manipulation platform that enhances horizontal force capability for push-and-pull interactions using two ducted fans integrated along the quadrotor interaction axis. An attitude-force decoupled control scheme enables controllable horizontal forces without requiring large attitude changes. Extensive real-world experiments are conducted to validate the DuctAM. Figure-eight trajectory tracking experiments demonstrate stable flight and accurate motion control in both quad and duct modes. Force-measurement experiments quantify the decoupled longitudinal force capability of DuctAM. Finally, representative push-and-pull interaction tasks, including cart pushing, door closing, and drawer opening, verify the practical effectiveness of DuctAM. The results show that DuctAM achieves significantly improved horizontal interaction force capability while maintaining stable flight compared with conventional UAVs.

\end{abstract}
\section{Introduction}
\label{sec:introduction}

Uncrewed Aerial Manipulators (UAMs) extend the capabilities of Uncrewed Aerial Vehicles (UAVs) from perception-centric autonomy~\cite{yao2026arsgaussian,zhang2024npe,zhang2025grounded,jin2024gs} to physical interaction with the environment~\cite{allenspach2025hri}.
Among various forms of aerial interaction, pushing and pulling represent fundamental manipulation primitives.
Many practical aerial tasks can be formulated as push-and-pull interactions, such as opening doors or drawers\cite{gupta2025umi} during search-and-rescue operations, pushing carts or movable containers\cite{deng25tro} in warehouse logistics, and pulling or guiding cables\cite{jin2025tethered} during tether base transportation.

\begin{figure}[!t]
    \begin{center}
        \includegraphics[width=1.0\columnwidth]{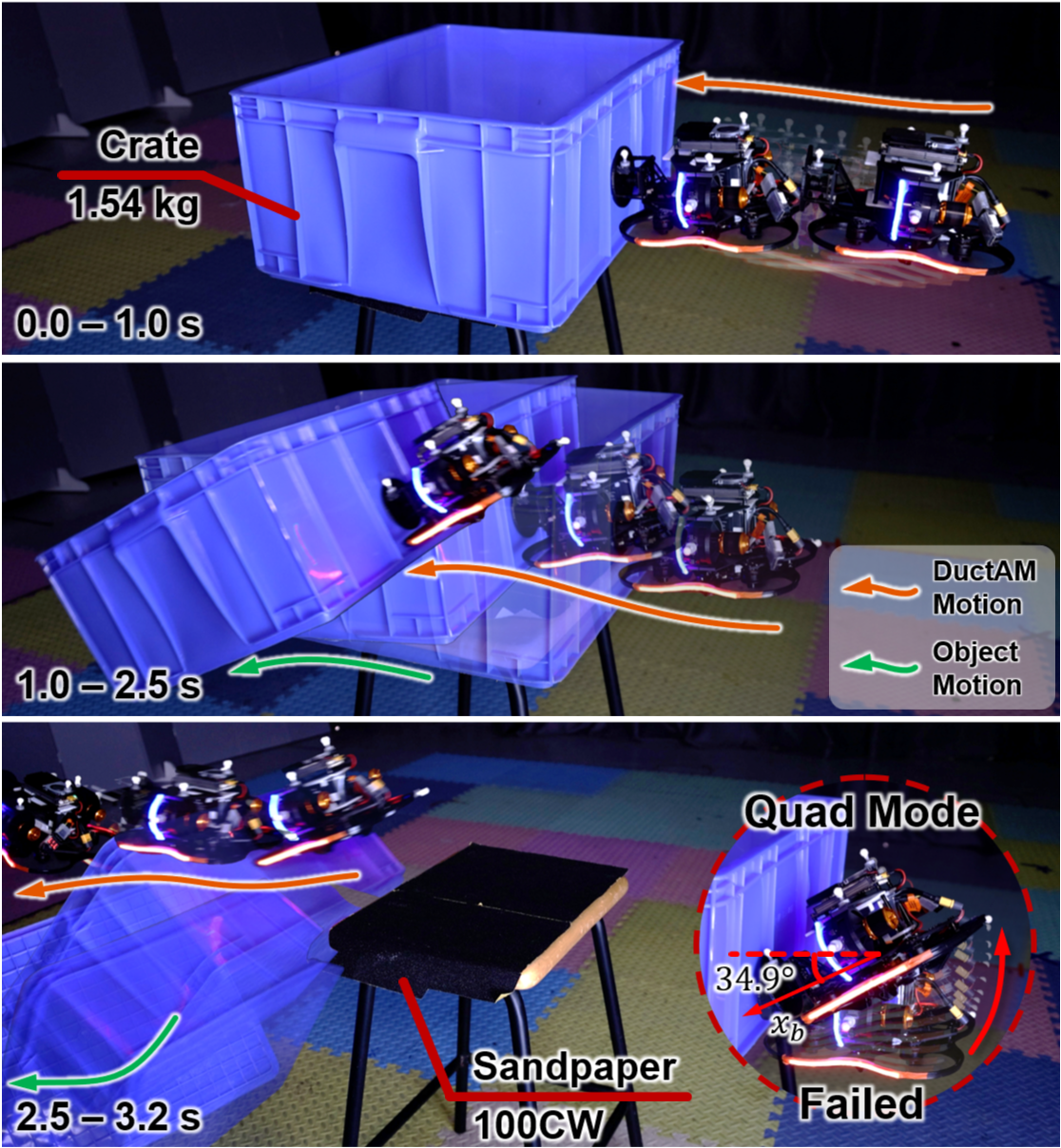}
    \end{center}
    \caption{DuctAM pushing a plastic crate on a sandpaper surface.
    }
    \label{fig:head}
\end{figure}

These tasks, such as the crate-pushing scenario illustrated in Fig.~\ref{fig:head}, require UAMs to generate sustained interaction forces while maintaining stable flight. Achieving reliable push-and-pull interactions therefore imposes the following key requirements on new generation UAMs:

\begin{itemize}

\item \textbf{High Lateral Force:}
Push-and-pull interactions require the UAMs to generate sufficiently large horizontal forces to overcome environmental resistance and drive object motion \cite{ollero2021past}.

\item \textbf{Attitude–Force Decoupling:}
The generated interaction forces should not significantly disturb the vehicle floating attitude, enabling stable flight and consistent contact direction during interaction.

\item \textbf{Structural Compactness:}
The aerial platform should maintain a compact structural footprint to operate effectively in confined or cluttered environments \cite{lin2025float}.

\end{itemize}

However, satisfying these requirements remains challenging for existing UAMs.
Conventional multi-rotor UAMs \cite{hongming25ICRA,zhaopeng25tmech,cao2025nature,meng25tro,sihao25quaduam,cao2024aircrab,wu2026hand,wu2023ring,li2026aerothrow} generate thrust primarily along the body normal direction.
Therefore, horizontal forces can only be produced by tilting the airframe, which introduces strong coupling between attitude and force generation.
As a result, large interaction forces may significantly disturb the UAM's attitude, limiting stability and controllability during interactions.

To overcome these limitations, several fully actuated or over-actuated aerial platforms
have been proposed to achieve six-DoF torque-force decoupling. However, these designs often introduce increased structural complexity and enlarged platform size. In addition, many fully actuated or thrust-vectoring designs employ tilted or spatially distributed propellers and vectoring mechanisms \cite{he2025flying,khj25ijrr,jinjie25teleoperate,kmj25pushing,park2018odar,kara2025design,isaac2025sensing}, whose thrust vectors are not always aligned with the desired interaction direction.

Consequently, only the projected component of the generated thrust contributes to the interaction force.
Moreover, a significant portion of the available thrust must be used to balance gravity and maintain flight, leaving a limited force margin for interaction.
As a result, the achievable force along a single horizontal direction remains constrained.
\begin{figure*}[!t]
    \begin{center}
        \includegraphics[width=2.0\columnwidth]{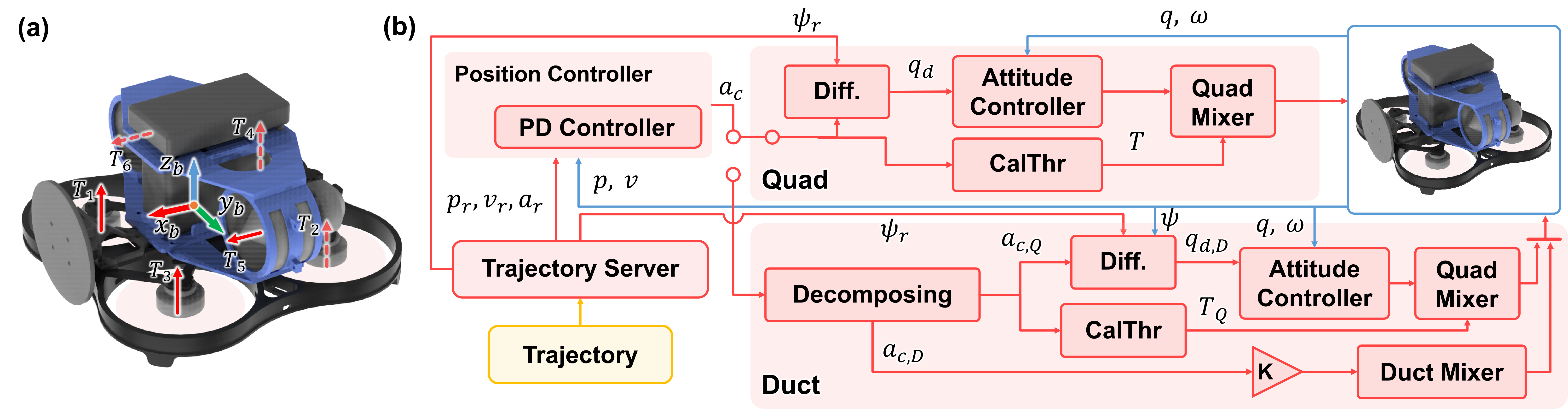}
    \end{center}
    \caption{Coordinate and force definitions, and the control framework.}
    \label{fig:2}
\end{figure*}

To address the challenges in horizontal push-and-pull interaction commonly encountered in aerial manipulation tasks, this paper proposes DuctAM, an aerial manipulation platform designed to enhance high horizontal force capability while maintaining structural compactness and stable flight.
The proposed architecture symmetrically integrates two ducted fans along the longitudinal axis of a quadrotor platform.
By leveraging counter-rotating motors and decoupled thrust control, the system can generate controllable horizontal forces without requiring large attitude changes.
This design significantly increases the available lateral force while preserving a compact structural footprint, thereby enabling high-force push-and-pull interactions in confined environments.

To validate the effectiveness of DuctAM for high-force push-and-pull interactions, we conduct extensive real-world experiments on representative aerial manipulation tasks, including drawer opening, door closing, cart pushing, and high-friction box pushing on sandpaper surfaces. These experiments prominently demonstrate the high-force interaction capabilities of DuctAM compared to conventional UAVs. Furthermore, a figure-eight trajectory attitude tracking experiment is conducted to exhibit the control performance and stability, particularly along the longitudinal axis.

    The main contributions of this paper are as follows:
\begin{itemize}
    \item We propose a novel UAM design, DuctAM, which is suitable for high-force push-and-pull interaction tasks. Through dual-ducted fans design, we achieve a platform with high longitudinal force capability.
    \item We introduce an acceleration decomposition scheme to overcome the coupling of attitude and force in conventional UAVs, thereby enabling attitude stabilization during sustained force exertion.
    \item Through a series of real-world experiments, we validate the capabilities of DuctAM in high-force push-pull manipulation scenarios.
\end{itemize}

\section{System Modeling}
\label{sec:modeling}
In this section, we formulate the dynamics of the proposed hybrid aerial platform, which integrates a standard quadrotor with two horizontal ducted fans. We define the world inertial frame as $\mathcal{F}_{\mathcal{W}} = \{x_w, y_w, z_w\}$ and the body-fixed frame as $\mathcal{F}_{\mathcal{B}} = \{x_b, y_b, z_b\}$, with the origin shown in Fig. 2(a). The system state is defined by position $\mathbf{p} \in \mathbb{R}^3$, velocity $\mathbf{v} \in \mathbb{R}^3$, attitude rotation matrix $\mathbf{R} \in SO(3)$, and body angular velocity $\boldsymbol{\omega} \in \mathbb{R}^3$.

\subsection{Hybrid Propulsion Model}

Unlike conventional multirotors, the proposed platform is equipped with two hybrid propulsion subsets: four vertically mounted rotors that generate thrusts denoted as $T_1$–$T_4$, and two horizontally mounted ducted fans that generate thrusts denoted as $T_5$ and $T_6$. The ducted fans are aligned parallel to the $x_b$-axis.

The total force $\mathbf{F}_b = [F_x, F_y, F_z]^T$ generated in the body frame can be decoupled into the quadrotor contribution and the ducted-fan contribution, defined as:

\begin{equation}
    \mathbf{F}_b = \mathbf{F}_{q} + \mathbf{F}_{d} =
    \begin{bmatrix}
        0 \\ 0 \\ \sum_{i=1}^4 T_i
    \end{bmatrix} +
    \begin{bmatrix}
        T_5 + T_6 \\ 0 \\ 0
    \end{bmatrix},
\end{equation}
where the subscripts $q$ and $d$ indicate the quadrotor and ducted-fan contributions, respectively. In the nominal model, the two ducted fans are assumed to be symmetrically commanded with $T_5=T_6$. Under this assumption, the ducted-fan subsystem provides only longitudinal push-and-pull force and does not contribute to yaw actuation. Similarly, the total control torque $\boldsymbol{\tau}_b = [\tau_x, \tau_y, \tau_z]^T$ in $\mathcal{F}_{\mathcal{B}}$ is given by
\begin{equation} \label{eq:torque}
    \boldsymbol{\tau}_b = \boldsymbol{\tau}_{q} + \boldsymbol{\tau}_{d}.
\end{equation}

The quadrotor torque is defined as:
\begin{equation}
    \boldsymbol{\tau}_{q} =
    \begin{bmatrix} \tau_{x,q} \\ \tau_{y,q} \\ \tau_{z,q} \end{bmatrix}
    =
    \begin{bmatrix}
        \frac{\sqrt{2}}{2} l_q (-T_1 + T_2 + T_3 - T_4) \\[0.3em]
        \frac{\sqrt{2}}{2} l_q (-T_1 - T_2 + T_3 + T_4) \\[0.3em]
        c_\tau (-T_1 + T_2 - T_3 + T_4)
    \end{bmatrix},
\end{equation}
where the effective lever arm and the drag torque coefficient of the vertical rotors are denoted by $l_q$ and $c_\tau$, respectively. The augmented torque from the ducted fans is explicitly expressed as:
\begin{equation}
    \boldsymbol{\tau}_{d} = \big[ 0, \ h_d F_{x,d}, \ 0 \big]^T,
\end{equation}
where $h_d$ represents the vertical offset of the ducted fans' thrust line from the CoM.

From the established propulsion model, two distinct characteristics of this hybrid configuration emerge. First, the ducted fans provide independent, fully decoupled control over the $x_b$-axis translation via the direct longitudinal actuation $F_{x,d}$, which fundamentally expands the underactuated flight envelope of the quadrotor. Second, while lateral parasitic forces are eliminated due to the parallel alignment, any vertical misalignment ($h_d \neq 0$) of the ducted fans introduces a coupled pitching moment $\tau_{y,d}$ during forward acceleration. This pitch coupling effect is treated as an internal disturbance and is actively rejected by the robust attitude controller.

\subsection{Rigid Body Dynamics}
The translational and rotational dynamics of the platform are governed by the Newton-Euler equations:
\begin{align}
    m\ddot{\mathbf{p}} &= m\mathbf{g} + \mathbf{R} \mathbf{F}_b \label{eq:trans_dyn},\\
    \mathbf{J}\dot{\boldsymbol{\omega}} &= \boldsymbol{\tau}_b - \boldsymbol{\omega} \times (\mathbf{J}\boldsymbol{\omega}), \label{eq:rot_dyn}
\end{align}
where $m$ is the total mass, and $\mathbf{J} \in \mathbb{R}^{3 \times 3}$ is the inertia tensor, $\mathbf{g} = [0, 0, -g]^T$ is the gravity vector.

To further illustrate the advantage of the hybrid design, we can expand the translational dynamics (Eq.~\ref{eq:trans_dyn}) using Z-Y-X Euler angles ($\psi, \theta, \phi$). In conventional quadrotors ($F_{x,d} = 0$), generating a global acceleration $\ddot{x}$ requires tight coupling with the pitch angle $\theta$. In contrast, for our platform operating in the duct mode, the pitch angle is constrained to zero ($\theta = 0$) for longitudinal stabilization when an interaction occurs. By substituting $\theta = 0$ into the rotation matrix $\mathbf{R}$, the translational equations simplify to an analytical form:
\begin{equation} \label{eq:decoupled_trans}
    m \begin{bmatrix} \ddot{x} \\ \ddot{y} \\ \ddot{z} + g \end{bmatrix} =
    \begin{bmatrix}
        F_{x,d}\cos\psi + F_{z,q}\sin\psi\sin\phi \\
        F_{x,d}\sin\psi - F_{z,q}\cos\psi\sin\phi \\
        F_{z,q}\cos\phi
    \end{bmatrix}.
\end{equation}

Eq.~\ref{eq:decoupled_trans} shows that the global longitudinal and lateral accelerations can be regulated by combining the horizontal force $F_{x,d}$ and the roll angle $\phi$ without tilting the airframe forward or backward. This strict pitch decoupling ensures minimal downwash airflow disturbance and lateral stabilization during close-proximity manipulations.

\begin{figure}[t]
    \centering
    \includegraphics[width=0.9\columnwidth]{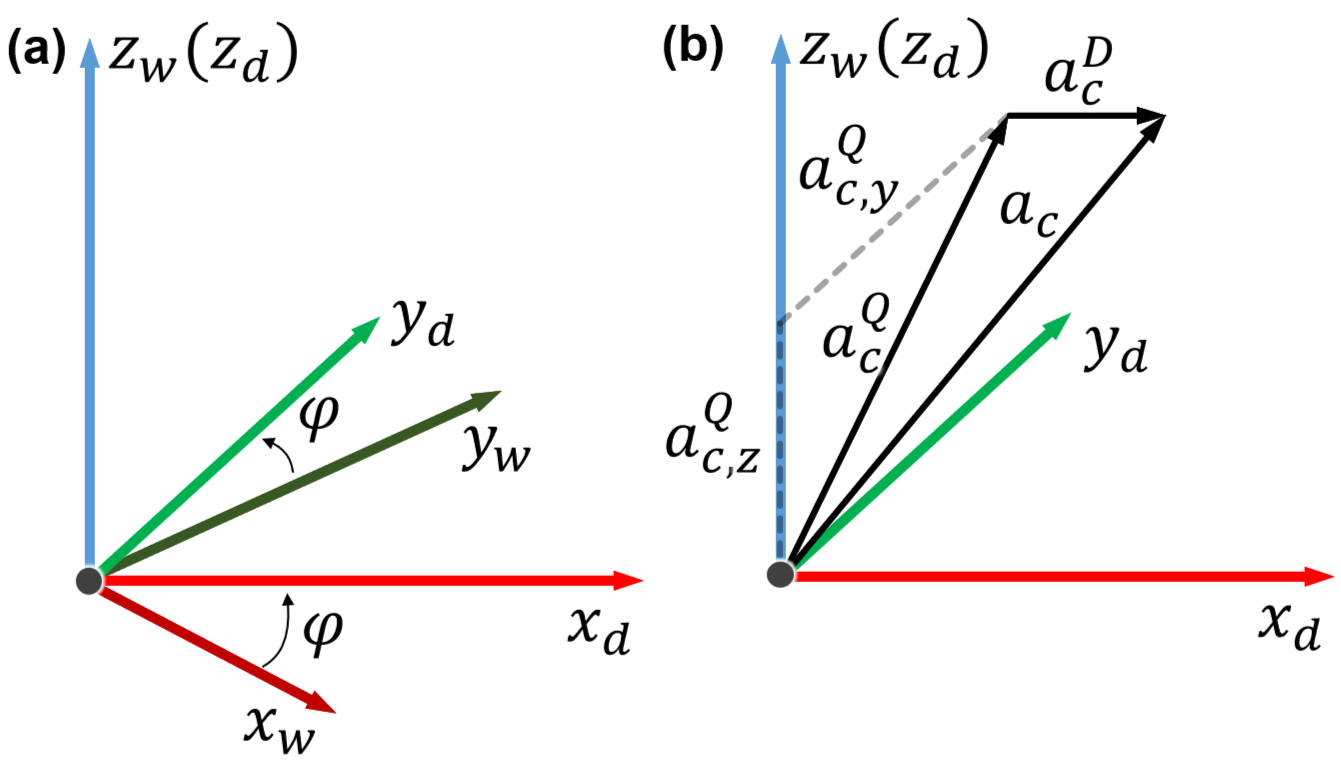}
    \caption{Acceleration decomposition in duct mode.}
    \label{fig:2_5}
\end{figure}

\section{Control Architecture}
\label{sec:control}

To exploit the platform's hybrid actuation, the control framework is designed to seamlessly switch between the underactuated quad mode and the decoupled duct mode. The framework is shown in Fig.~\ref{fig:2}(b) and primarily consists of two parts: the position controller and the attitude controller.

\subsection{Position Control}
The position controller is implemented as a cascaded PD controller. First, the desired global acceleration $\ddot{\mathbf{p}}_d$ is calculated as follows:
\begin{equation}
    \ddot{\mathbf{p}}_d = \ddot{\mathbf{p}}_r + \mathbf{K}_v(\dot{\mathbf{p}}_r + \mathbf{K}_p(\mathbf{p}_r - \mathbf{p}) - \dot{\mathbf{p}}),
\end{equation}
where $\ddot{\mathbf{p}}_r$, $\dot{\mathbf{p}}_r$, and $\mathbf{p}_r$ denote the reference acceleration, velocity, and position obtained from the reference trajectory; $\dot{\mathbf{p}}$ and $\mathbf{p}$ represent the measured velocity and position of the drone; and $\mathbf{K}_v, \mathbf{K}_p$ are positive definite gain matrices.

When the drone operates in the underactuated quad mode, it can only generate thrust along its body Z-axis $\mathbf{z}_{\mathcal{B}}$. The desired collective thrust $T_{z,d}$ is expressed as:
\begin{equation}
    T_{z,d} = (m\ddot{\mathbf{p}}_d - m\mathbf{g}) \cdot \mathbf{z}_{\mathcal{B}}.
\end{equation}

\begin{figure*}[!t]
    \begin{center}
        \includegraphics[width=1.8\columnwidth]{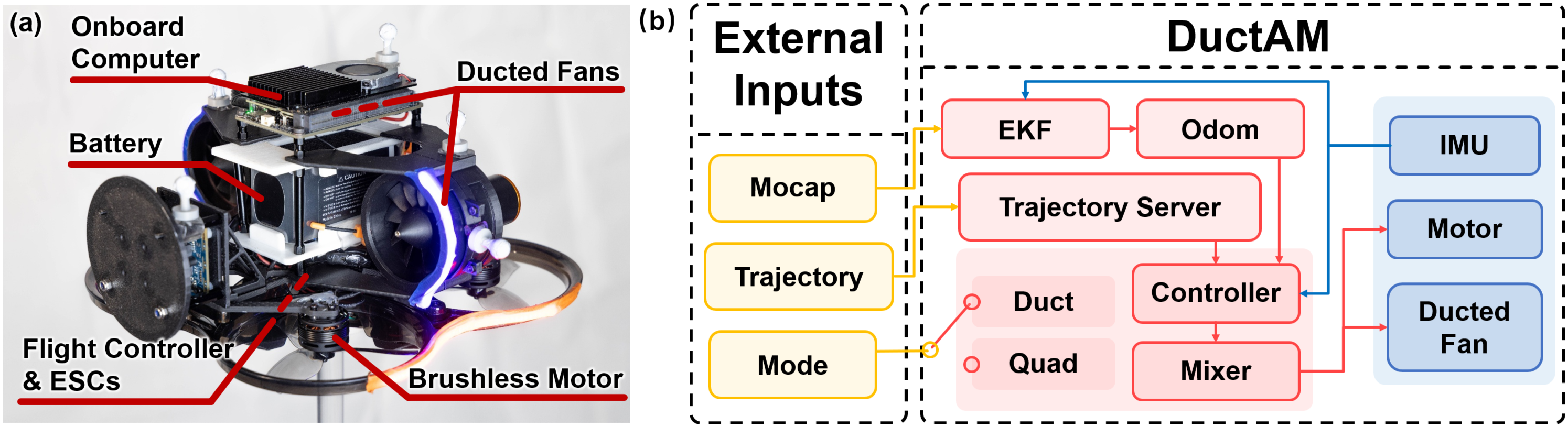}
    \end{center}
    \caption{Hardware configuration and system overview.}
    \label{fig:3}
\end{figure*}

When switching to the duct mode, the drone achieves decoupled longitudinal
control. The execution along the horizontal $X$-axis is shifted to the
ducted fans, while the position error is still corrected by the controller.
Therefore, the desired force can be expressed in the body frame as
\begin{equation}
    \mathbf{f}_{b,d}
    =
    \mathbf{R}^{T}
    \left(
    m\ddot{\mathbf{p}}_d - m\mathbf{g}
    \right)
    =
    \begin{bmatrix}
        f_{x,d} \\
        f_{y,d} \\
        f_{z,d}
    \end{bmatrix},
\end{equation}
where $\mathbf{f}_{b,d}$ denotes the desired body-frame force vector. The
longitudinal component $f_{x,d}$ is assigned to the ducted fans, while the
vertical component $f_{z,d}$ is mainly provided by the quadrotor subsystem.

\subsection{Attitude Control and Decoupling}
The attitude controller is formulated as a cascaded PID controller operating on the $SO(3)$ manifold. First, we define the attitude error $\mathbf{R}_e$ as:
\begin{equation}
    \mathbf{R}_e = \frac{1}{2}(\mathbf{R}_r^T \mathbf{R} - \mathbf{R}^T \mathbf{R}_r)^\vee,
\end{equation}
where $\mathbf{R}_r$ and $\mathbf{R}$ denote the reference and measured attitude rotation matrices, respectively, and $(\cdot)^\vee$ represents the mapping of elements from the Lie algebra $\mathfrak{so}(3)$ to vectors in $\mathbb{R}^3$.

For the quad mode, $\mathbf{R}_r$ is derived from the desired acceleration $\ddot{\mathbf{p}}_d$ and the reference yaw using the differential flatness property. However, in the duct mode, the forward acceleration is directly provided by the ducted fans. To prevent this independent thrust from coupling with the differential flatness mapping, the total desired acceleration is required to be decomposed.

Specifically, as illustrated in Fig. \ref{fig:2_5}(a), we introduce a yaw-aligned intermediate frame $\mathcal{D}$ (comprising axes $x_d, y_d, z_d$). This frame is obtained by rotating the world frame $\mathcal{W}$ around the $z_w$-axis by the reference yaw angle $\psi_r$. The total desired acceleration vector $\bm{a}_{c}$ is then decomposed in Fig. \ref{fig:2_5}(b) as:
\begin{equation}
\bm{a}_{c} = \bm{a}_{c}^{Q} + \bm{a}_{c}^{D},
\end{equation}
where $\bm{a}_{c}^{Q}$ is the acceleration distributed to the quadrotor subsystem, and $\bm{a}_{c}^{D}$ is the acceleration distributed to the ducted fans. Since the ducted fans' control authority is constrained to the $x_d$-axis, the scalar acceleration allocated to the ducted fans, denoted as $a_{\rm duct}$, is obtained by orthogonally projecting $\bm{a}_c$ onto the $x_d$-axis of the intermediate frame $\mathcal{D}$:
\begin{equation}
  a_{duct} = \mathbf{e}_1^T \mathbf{R}_z(\psi_r)^T \bm{a}_c,
\end{equation}where $\mathbf{e}_1 = [1, 0, 0]^T$ and $\mathbf{R}_z(\psi_r) \in SO(3)$ is the rotation matrix from $\mathcal{D}$ to $\mathcal{W}$. The spatial acceleration vector allocated to the ducted fans in the world frame is then formulated as:
\begin{equation}
    \bm{a}_c^D = \mathbf{R}_z(\psi_r) \begin{bmatrix} a_{duct} \\ 0 \\ 0 \end{bmatrix}.
\end{equation}

After isolating the ducted fan's effort, the residual acceleration $\bm{a}_c^Q = \bm{a}_c - \bm{a}_c^D$ accounts entirely for the vertical lift and lateral maneuvering. The decoupled quadrotor acceleration $\bm{a}_c^Q$ and the reference yaw $\psi_r$ are utilized through the standard differential flatness mapping to generate the decoupled reference attitude matrix $\mathbf{R}_r$.

This acceleration decomposition naturally drives the flatness mapping to yield a near-zero pitch angle ($\theta \approx 0$), effectively decoupling the longitudinal translation from the pitch dynamics while strictly tracking the altitude and lateral trajectories.

The desired angular velocity $\boldsymbol{\omega}_d$ and angular velocity error $\boldsymbol{\omega}_e$ are defined as follows:
\begin{align}
    \boldsymbol{\omega}_d &= \mathbf{K}_R \cdot \mathbf{R}_e, \\
    \boldsymbol{\omega}_e &= \boldsymbol{\omega}_d - \boldsymbol{\omega},
\end{align}
where $\mathbf{K}_R$ is a positive definite gain matrix, and $\boldsymbol{\omega}$ represents the measured angular velocity. Then, the desired torque $\boldsymbol{\tau}_{des}$ is calculated using positive definite PID gain matrices $\{\mathbf{K}_{P,\omega}, \mathbf{K}_{I,\omega}, \mathbf{K}_{D,\omega}\}$:
\begin{equation}
    \boldsymbol{\tau}_{des} = \mathbf{K}_{P,\omega}\boldsymbol{\omega}_e + \mathbf{K}_{I,\omega}\int \boldsymbol{\omega}_e dt + \mathbf{K}_{D,\omega}\dot{\boldsymbol{\omega}}_e.
\end{equation}

For control allocation, the computed desired torque $\boldsymbol{\tau}_{des}$ and the collective thrust derived from $\bm{a}_c^Q$ are allocated through the quadrotor mixer. Simultaneously, in the duct mode, the decoupled longitudinal acceleration
scalar $a_{\rm duct}$ is mapped to the ducted fan actuators. Since the two
ducted fans are symmetrically arranged and commanded with equal magnitude,
the duct mixer generates the actuator commands as
\begin{equation}
    \begin{bmatrix}
        u_5 \\
        u_6
    \end{bmatrix}
    =
    \mathbf{K}_{\rm duct} a_{\rm duct},
\end{equation}
where $\mathbf{K}_{\rm duct} = [k_5, k_6]^T$ is the mapping gain vector from
the desired longitudinal acceleration to the ducted-fan commands. In this
work, $k_5=k_6$ is used to avoid introducing an additional yaw moment.

\begin{table}[t]
	\footnotesize
	\renewcommand{\arraystretch}{1.1}
	\setlength{\tabcolsep}{3pt}
	\centering
	\caption{Component Configuration and Weight Distribution}
	\label{tab:component_weights}
	\noindent
	\resizebox{\linewidth}{!}{
		\begin{tabular}{@{}
				>{\raggedright\arraybackslash}p{2.2cm} |
				>{\raggedright\arraybackslash}p{3.3cm} |
				c  |
				r
				@{}}
			\toprule[2pt]
			\textbf{Component} &
			\textbf{Model} &
			\textbf{Units} &
			\textbf{Weight (g)}\\
			\midrule
                Battery           & VOLTEN 1550\,mAh 6S      & 1 & 233 \\
                \mbox{Onboard Computer}  & NVIDIA Jetson Orin NX              & 1 & 81  \\
                Ducted Fan        & QX-MOTOR 50\,mm        & 2 & 146 \\
                Brushless Motor   & \mbox{2004.5 2250\,KV Customized}  & 4 & 73.6\\
                ESC               & HAKRC 65A                          & 2 & 22  \\
                Flight Controller & NxtPX4v2                           & 1 & 6.5 \\
			\bottomrule[2pt]
	\end{tabular}}
	\vspace{-1.0cm}
\end{table}

\begin{figure*}[!t]
    \begin{center}
        \includegraphics[width=2.0\columnwidth]{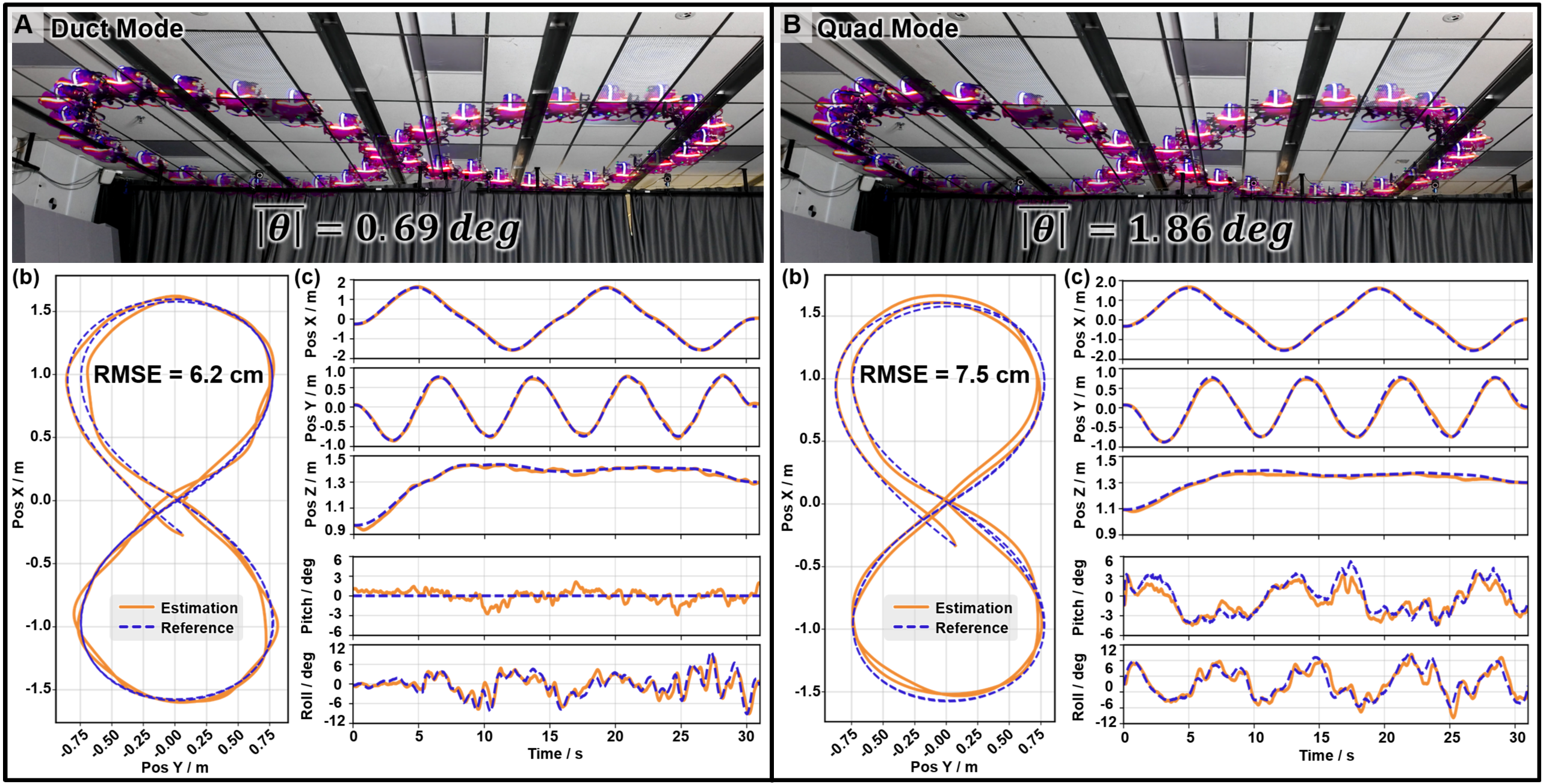}
    \end{center}
    \caption{Figure-8 trajectory tracking experiments and results. Subfigures A and B show the experimental results of DuctAM in duct mode and quad mode, respectively: (a) trajectory snapshot overlay, (b) position tracking curves, and (c) time-series of the drone position and attitude during trajectory tracking.}
    \label{fig:8_traj}
\end{figure*}

\section{Experiments}
\subsection{Implementation Details}
We built a prototype DuctAM to validate the proposed dynamics and control framework. The hardware configuration is shown in Fig.~\ref{fig:3}(a). The prototype measures $\ 230\times220\times130\ \mathrm{mm}$ and weighs$\ 987\ \mathrm{g}$. It consists of a conventional quadrotor base with two ducted fans mounted in parallel along the body longitudinal axis. The ducted fans are positioned on a plane $\ 1\ \mathrm{cm}$ above the center of mass. DuctAM integrates an NVIDIA Jetson Orin NX as the onboard computer and an NxtPX4v2 flight controller for low-level attitude stabilization, motor command execution, and onboard IMU data acquisition. Table~\ref{tab:component_weights} summarizes the model and weight distribution of key components.

The system overview is illustrated in Fig.~\ref{fig:3}(b). State estimation is provided by an Extended Kalman Filter (EKF) that fuses measurements from a Vicon motion capture system and the onboard IMU, producing odometry for the controller. The prototype can track a commanded reference trajectory using the proposed control pipeline.

\subsection{Figure-8 Trajectory Tracking}
To evaluate the control performance and tracking accuracy of the proposed platform, we conducted figure-8 trajectory tracking experiments on DuctAM. The platform was commanded to track a $3.2 \times 1.6~\mathrm{m}$ figure-8 trajectory in the quad mode and the duct mode, respectively. In each trial, DuctAM completed two laps along the reference path. The experimental results are visualized in Fig.~\ref{fig:8_traj}.

To further quantify the longitudinal force--pitch decoupling performance of DuctAM, we introduce the mean absolute pitch angle as an evaluation metric, defined as
\begin{equation}
\overline{|\theta|}=\frac{1}{N}\sum_{i=1}^{N}\left|\theta_i\right|,
\label{eq:mean_abs_pitch}
\end{equation}
where $N$ is the total number of samples and $\theta_i$ denotes the pitch angle at the $i$-th sample.
From the experimental data, we obtain $\overline{|\theta|}=0.69^{\circ}$ in the duct mode and $\overline{|\theta|}=1.86^{\circ}$ in the quad mode, indicating that DuctAM can generate longitudinal force with substantially reduced pitch deviation, thus validating the decoupling effect.

Meanwhile, we evaluate the 2D position tracking accuracy in the $x$--$y$ plane using the root-mean-square error (RMSE). The position RMSE is $6.2~\mathrm{cm}$ in the duct mode and $7.5~\mathrm{cm}$ in the quad mode, indicating comparable tracking accuracy and the potential of DuctAM for precise motion tracking with enhanced longitudinal force generation.

\subsection{Longitudinal Force Measurement}

\begin{figure*}[!t]
    \begin{center}
        \includegraphics[width=2.0\columnwidth]{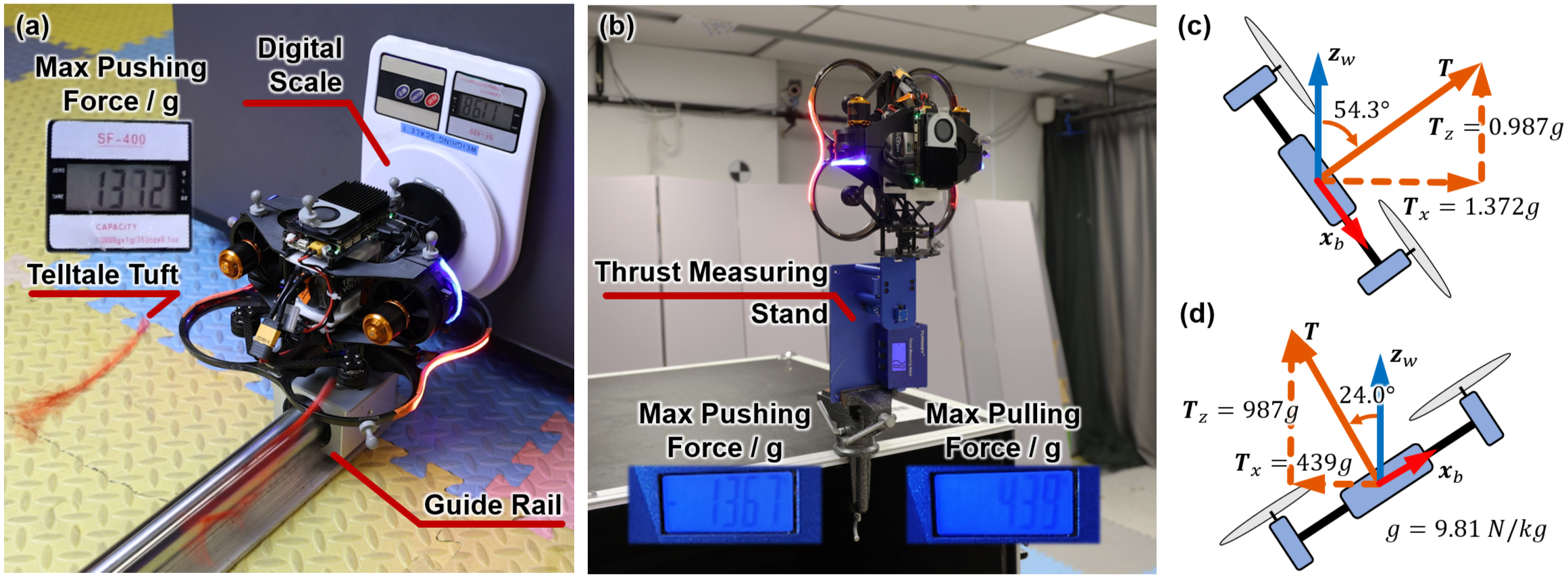}
    \end{center}
    \caption{Longitudinal force measurement. (a) A force-measurement rig based on a digital scale. (b) A force-measurement rig based on a thrust-measuring stand. (c) and (d) Required pitch angles in the quad mode to generate longitudinal force components equal to DuctAM's maximum pushing and pulling forces, respectively.}
    \label{fig:force}
\end{figure*}

A key advantage of DuctAM over a conventional quadrotor is that the longitudinal force can be regulated independently of the pitch angle, which makes it suitable for push-and-pull interactions. To quantify the decoupled longitudinal force that DuctAM can generate, we performed the experiment shown in Fig.~\ref{fig:force}.

To ensure measurement accuracy, we built two test rigs and cross-validated the results, as shown in Fig.~\ref{fig:force}(a) and (b). In Fig.~\ref{fig:force}(a), the force measurement is based on a digital scale: the quadrotor base is rigidly mounted on a linear guide rail to constrain the motion to the longitudinal direction, and the front head of DuctAM is brought into contact with the scale to measure the interaction force. Using this setup, we measured a maximum pushing force of $1372~\mathrm{g}$.

However, this rig is not suitable for measuring pulling force because the ducted-fan airflow impinges on the scale and leads to biased readings. Therefore, we further built a thrust-measuring-stand-based rig, where the drone is mounted vertically on the stand shown in Fig.~\ref{fig:force}(b). This rig introduces negligible aerodynamic interference, and the measured maximum pushing and pulling forces are $1367~\mathrm{g}$ and $439~\mathrm{g}$, respectively. The maximum pushing force differs from the digital-scale measurement by only $0.37\%$, supporting the reliability of the measurements.

For comparison with a conventional quadrotor, we compute the pitch angles required in the quad mode to generate the same longitudinal pushing/pulling forces while maintaining vertical force balance against gravity, as shown in Fig.~\ref{fig:force}(c) and (d). The quad mode requires pitch angles of $54.3^{\circ}$ and $-24.0^{\circ}$ to produce the corresponding maximum pushing and pulling forces, respectively. These results highlight the force--attitude decoupling capability of DuctAM and demonstrate the potential of the proposed configuration for high-force interaction tasks with reduced attitude deviation.

\subsection{Push-and-Pull Interactions}
To evaluate the practicality of DuctAM in push-and-pull interactions, we conducted four representative tasks: cart pushing, door closing, and drawer opening (Fig.~\ref{fig:tasks}), as well as crate pushing on sandpaper (Fig.~\ref{fig:head}).

\subsubsection{Cart Pushing}
A key advantage of the ducted-fan configuration is its ability to provide a large thrust magnitude, which enables DuctAM to perform pushing interaction tasks with relatively heavy loads. To validate this capability, we conducted a cart-pushing experiment, as shown in Fig.~\ref{fig:tasks}.A. In this experiment, DuctAM was commanded to push a cart on a foam crawling mat, with the cart carrying a $3.87~\mathrm{kg}$ payload (a wheel and a plastic crate).

DuctAM maintained continuous contact for $6.2~\mathrm{s}$ and pushed the cart forward by approximately $2~\mathrm{m}$. During pushing, DuctAM maintained a nearly level attitude, with the pitch angle staying around $1^{\circ}$. We also tested the drone in the quad mode under the same conditions. The quadrotor failed to move the cart and became unstable, even when the pitch angle exceeded $21.2^{\circ}$. This loaded cart-pushing experiment demonstrates the ability of DuctAM to generate longitudinal thrust for pushing interactions.

\subsubsection{Door Closing}
To evaluate the performance of DuctAM in pushing a rotating object, we conducted a door-closing experiment that mimics a common daily action, as shown in Fig.~\ref{fig:tasks}(b). DuctAM was tasked with closing an aviation-case door of size $0.88 \times 1.06~\mathrm{m}$.

DuctAM contacted the door at $t=2.4~\mathrm{s}$ and pushed continuously for about $3.1~\mathrm{s}$ until it was fully closed. During the interaction, DuctAM maintained a consistent yaw rotation, with a total yaw change exceeding $30^{\circ}$.

\subsubsection{Drawer Opening}
\begin{figure*}[!t]
    \begin{center}
        \includegraphics[width=2.0\columnwidth]{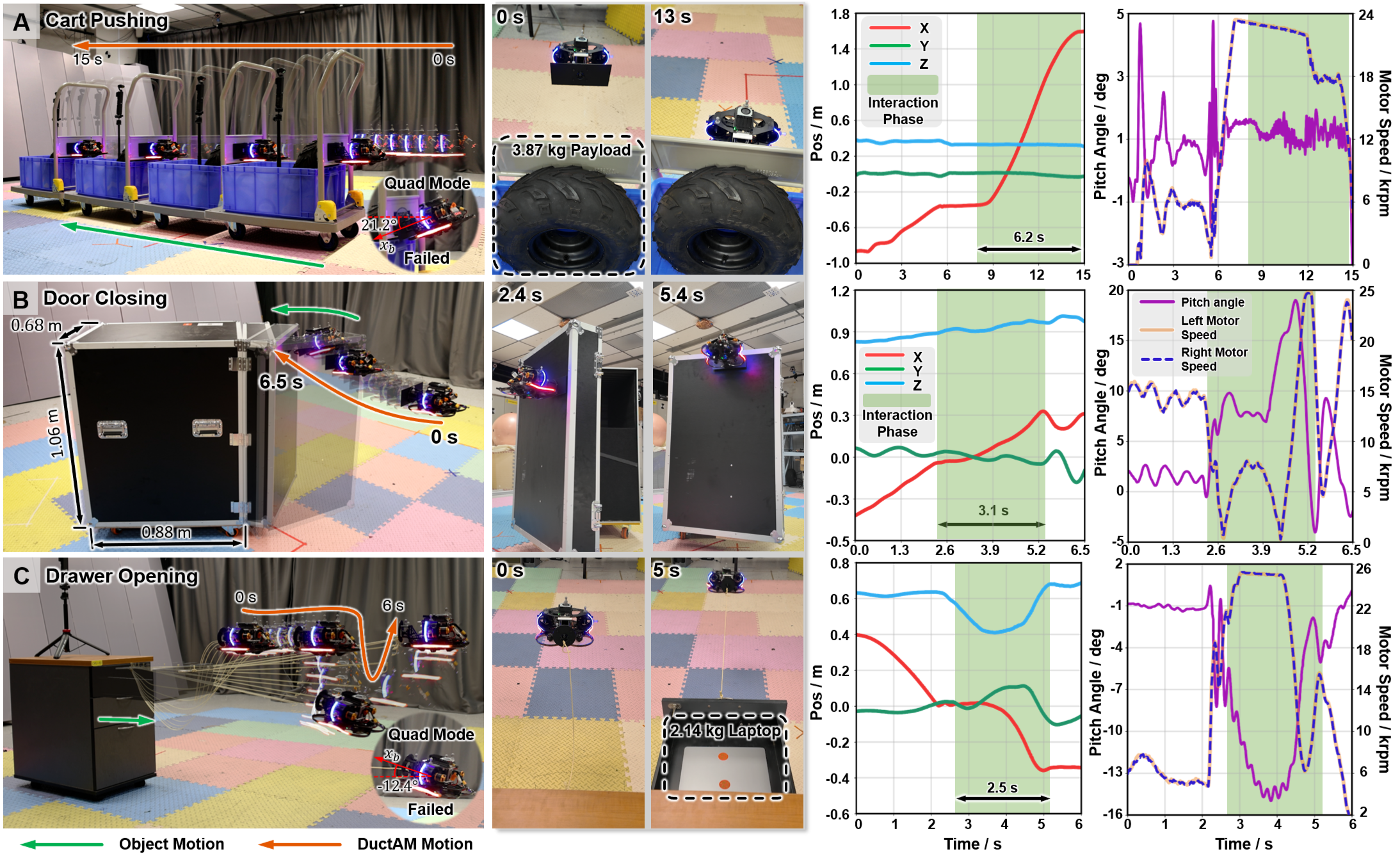}
    \end{center}
    \caption{Push-and-pull tasks. Each row corresponds to a different task. From left to right, the columns show a snapshot of the interaction, external-view images, the estimated position curves, the pitch-angle profile, and the ducted-fan RPM profile.}
    \label{fig:tasks}
\end{figure*}

To verify the feasibility of DuctAM for pulling interaction tasks, we designed a daily-life scenario: drawer opening (Fig.~\ref{fig:tasks}.C). A $2.14~\mathrm{kg}$ laptop was placed inside a closed drawer, and a rope was attached between the drawer handle and the front end of DuctAM. By reversing the thrust direction of the ducted fans, DuctAM generated the pulling force and successfully opened the drawer after approximately $2.5~\mathrm{s}$ of continuous pulling, during which the rope remained taut.

For comparison, we also evaluated the performance of DuctAM in the quad mode under the same setup as a benchmark. The quadrotor failed to open the drawer even when its pitch angle reached $-12.4^{\circ}$. Despite exhibiting a similar pitch angle to the quad mode, the configuration of DuctAM provides additional horizontal thrust, enabling the drawer to be opened without requiring large attitude changes. These results demonstrate the capability of the proposed DuctAM system for pulling interactions.

\subsection{Stability Under Contact Force Loss}
During aerial interaction, a sudden loss of contact with the object poses a significant stability challenge. To validate this capability, we designed an experiment in which DuctAM pushes and topples a plastic crate placed on a 100CW sandpaper surface. As shown in Fig.~\ref{fig:head}, the trial lasted $3.2~\mathrm{s}$. From $t=0$ to $1~\mathrm{s}$, DuctAM approached the crate and then continuously pushed for approximately $1.5~\mathrm{s}$ until the crate was toppled. After separation, DuctAM remained stable despite the abrupt release of interaction force.

For comparison, we also tested a conventional quadrotor configuration, which failed to push the crate even with the pitch angle increased to $34.9^{\circ}$. These results demonstrate the dynamic response and stability of DuctAM under sudden contact transitions, indicating its potential for robust high-force interaction tasks.

\section{Conclusion and Future Work}
In this work, we presented DuctAM, a compact aerial manipulation platform designed to enable high-force push-and-pull interactions while maintaining stable flight. By integrating two ducted fans along the longitudinal axis of a quadrotor and employing an attitude-force decoupled control strategy, the proposed system can generate enough longitudinal forces to accomplish practical tasks without requiring large attitude deviations.

To validate the effectiveness of DuctAM, extensive real-world experiments were conducted, including trajectory tracking, force measurement, and representative push-and-pull interaction tasks such as cart pushing, door closing, and drawer opening. The results demonstrate that DuctAM can produce significantly larger horizontal interaction forces while preserving stable flight performance compared with conventional UAV platforms.

These results highlight the potential of the proposed architecture for aerial manipulation tasks that require strong horizontal interaction force. Future work will investigate improved control strategies for contact-rich manipulation and extend the platform to more complex aerial interaction scenarios.

\bibliography{ICRA2022}
\end{document}